\documentclass[letterpaper, 10 pt, conference]{ieeeconf}  

\IEEEoverridecommandlockouts                              

\usepackage{graphics} 
\usepackage{epsfig} 
\usepackage{mathptmx} 
\usepackage{times} 
\usepackage{amsmath} 
\usepackage{amssymb}  
\usepackage{algorithm}
\usepackage{algpseudocode}
\usepackage{booktabs} 
\usepackage{tabularx}
\usepackage{graphicx}     
\usepackage{caption}      
\usepackage{subcaption}   
\usepackage{float}        
\usepackage{cite}

\title{\LARGE \bf
Text2VR: Automated instruction Generation in Virtual Reality using Large language Models for Assembly Task
}

\author{Subin Raj Peter$^{1}$ 
\thanks{*This work was not supported by any organization}
\thanks{$^{1}$Author is a researcher from University College Dublin, Ireland.
        {\tt\small subin.raj@ucd.ie}}%
}

\begin{document}

\maketitle
\thispagestyle{empty}
\pagestyle{empty}

\begin{abstract}

Virtual Reality (VR) has emerged as a powerful tool for workforce training, offering immersive, interactive, and risk-free environments that enhance skill acquisition, decision-making, and confidence. Despite its advantages, developing VR applications for training remains a significant challenge due to the time, expertise, and resources required to create accurate and engaging instructional content. To address these limitations, this paper proposes a novel approach that leverages Large Language Models (LLMs) to automate the generation of virtual instructions from textual input. The system comprises two core components: an LLM module that extracts task-relevant information from the text, and an intelligent module that transforms this information into animated demonstrations and visual cues within a VR environment. The intelligent module receives input from the LLM module and interprets the extracted information. Based on this, an instruction generator creates training content using relevant data from a database. The instruction generator generates the instruction by changing the color of virtual objects and creating animations to illustrate tasks. This approach enhances training effectiveness and reduces development overhead, making VR-based training more scalable and adaptable to evolving industrial needs.  
\end{abstract}

\section{INTRODUCTION}
The rapid development of technologies such as automation, digital marketing, artificial intelligence, and robotics is reshaping the nature of work, leading to a growing demand for new skill sets. Consequently, continuous training and upskilling of workers have become essential. Various methods are used to train workers in industrial environments, including expert-led instruction, printed manuals, videos, Augmented Reality (AR), and Virtual Reality (VR). Each method has its own strengths and limitations. In industrial settings, operators typically learn tasks with the assistance of experts through face-to-face training or external tools such as paper-based and video-based documents. However, document-based learning lacks realism, and the limited availability of experts makes training less accessible to operators \cite{singhaphandu2024manual}. VR has provided a flexible and immersive platform for training workers, with applications across various industries such as manufacturing \cite{tusher2024systematic}, education \cite{tusher2024systematic}, healthcare \cite{xie2021review}, the military \cite{leite2025use}, and construction \cite{leite2025use}. In manufacturing, VR has been used in various areas including training, factory layout planning, virtual prototyping, robot path planning, etc. Among the different training methods, VR has been widely used for training the worker in industries \cite{chandra2021implementation}.VR enables users to learn tasks in a simulated environment, offering greater flexibility in delivering instructions. Training content can be conveyed through multiple modalities, including text, audio, visual, and haptic feedback \cite{chuang2023visuo}. However, generating these instructions requires expert knowledge and specialized skills, and different applications must often be developed for each task. The development of VR applications is typically time-consuming and complex, which limits their scalability and broader adoption. Text is a common and accessible mode of communication. However, relying solely on textual instructions can be problematic \cite{gattullo2020and}. Text often lacks the detail which leads to ambiguity, misunderstandings, and errors \cite{kolla2021comparing}.

Natural Language Processing (NLP) algorithms offer a promising solution for extracting relevant information from instructional text. LLMs, a sophisticated application of NLP, have shown strong performance in tasks such as text-to-image generation \cite{xu2024llm} and text summarization \cite{zhang2023auto}. Recent evaluations demonstrate that LLMs can effectively extract task-relevant information from text. This work aims to leverage LLMs to generate virtual instructions from text input. The proposed method extracts the relevant information from the text, which is then used to generate the instruction. This approach introduces a novel method for transforming textual instructions into immersive VR training content. In addition to streamlining the content creation process, it enhances the effectiveness of training by providing intuitive, real-time guidance.Currently, many LLMs are openly available, such as ChatGPT \cite{chatgpt2025} and LLaMA \cite{llama2025}, offering easy access and usability. These models are trained on diverse datasets, enabling them to perform across a wide range of domains. However, their performance may decline when applied to specific domains due to a lack of domain-specific knowledge. This limitation can be addressed by fine-tuning the models for targeted applications. This work discusses the training of LLMs for specific domains and the generation of virtual instructions from text input. The main contributions of this work are summarized as follows:

\begin{enumerate}
    \item Proposed a pipeline and method for generating virtual instructions from textual input.
    \item Developed a intelligent module to generate the virtual instruction in the VR environment.
    \item Validated the proposed method through an actual assembly task, demonstrating the effectiveness of LLM-generated virtual instructions.
\end{enumerate}

The organization of the paper is as follows. We discussed the current state of the art of VR based training and LLM in section 2. We explained the proposed method to generate virtual instruction from the text in section 3. Details of the evaluation is elaborated in section 4 followed by concluding remarks at sections 5.


\section{Literature Survey}
VR has emerged as one of the most effective tools for worker training due to its immersive and interactive capabilities \cite{daling2023assemble}. VR-based safety training significantly enhanced workers’ safety skills, auditing capabilities, and proactive risk management compared to traditional methods \cite{al2025virtual}. It also enables a better understanding of tasks and improves learning outcomes compared to PC-based and conventional training methods \cite{checa2023immersive}. VR allows users to learn tasks by doing, rather than just seeing, observing, and listening \cite{abidi2019assessment}. It is widely used in industries because it enables users to practice tasks away from the actual worksite, providing a realistic, immersive, and interactive environment. This allows workers to learn dangerous tasks without any risk \cite{brunzini2021virtual}. VR training modules are used to simulate complex manufacturing processes, motivate workers to learn, and facilitate collaboration among workers from different locations at reduced cost and time \cite{ipsita2025virtual}. VR based training has significant advantage over conventional training methods in terms of flexibility, time saving, and material production \cite{schwarz2020learning}. Additionally, VR enables groups of workers to acquire skills more quickly and effectively. Besides, VR based training limited by large amount of time and cost, user engagement, and error management \cite{eswaran2022challenges, studer2024open}. 

In VR, training information provided to the operator using different strategies including programmed instruction and gamification. Gamification based training reduce the operator error in the training process compared to non gamification based training \cite{buldu2025cuify}. Training in the virtual environment is cheaper, faster, and easier than real world training and it enables quick changes between different training scenarios \cite{niedermayr2021design}. The main function of the VR in the training process is to 1) provide accurate, real time guidance to the trainee with clear instructions; 2) identify the training step and confirm it performs correctly; 3) could able to update VR content upon special request from the trainee \cite{ho2018virtual}. In VR, instructions are provided to workers in the virtual environment through multiple modalities such as visual, audio, and haptic means \cite{ulmer2022gamification, wolfartsberger2023analyzing, niedermayr2021design}. This multimodal information transfer enhance the operator leaning with a detailed explanation. Training the worker using multiple modalities enhances knowledge transfer compared to using a single modality. These instructions are provided to the worker as per the request. This can be achieved by two main components of VR systems: the interaction layer and the control layer. The interaction layer assists users in interacting with the VR environment through their senses and accessing the instructions. The control layer manages the physics and dynamics of the components in the VR environment and controls the information transfer \cite{lan2023virtual}. Adding feedback about the training task can improve training effectiveness. To achieve this, tasks performed by users in the virtual environment need to be monitored. The feedback information can be created by continuously monitoring the components in the VR and user senses. The monitoring task can be further improved by using Convolutional Neural Networks (CNNs) to track human activity \cite{eversberg2021industrial}. VR allows trainee and trainer to share the same environment. This allows the trainer to monitor the training and provide necessary assistance during training process \cite{niedermayr2021design}. In VR based training methods, the instructions are created in advance by experts, and small changes in the task lead to redevelopment of the application. Introducing AI in the VR based training enables training without supervision of the trainer \cite{ho2018virtual}. LLM models can also used for generation environment aware interaction and conversation to improve interaction in VR \cite{li2025exploring}. The performance of VR-based training can be further enhanced by integrating advanced AI algorithms, which help reduce the time, complexity, and computational cost of training systems for hazardous tasks \cite{naranjo2020scoping}. 


LLM has been widely used as a chatbox where the operator chat with the LLM to retrieve the information. This information transferred to the operator as a audio signal and virtual character with lipsync \cite{li2025generative, buldu2025cuify}. LLMs have shown remarkable versatility across different fields of manufacturing, enhancing productivity and creativity in unique ways. In the manufacturing sector, LLMs facilitate collaborative design by providing visual interactive prompts that assist multiple users in the design process \cite{xu2024llm}. Additionally, LLMs have proven effective in code comprehension and generation, streamlining software development tasks \cite{yuan2023evaluating}. LLMs enable intuitive scene manipulation by extracting information from text. This information is used to modify the position and orientation of virtual objects \cite{wang2025can}. LLM-based system is presented for real-time prompting in interactive worlds, allowing users to generate detailed scenes from prompts, thus enhancing immersive virtual experiences \cite{de2024llmr}. This assists users in developing the virtual environment from text messages, simplifying the development of the VR environment. For training workers, instructions need to be developed effectively inside the virtual environment to assist users. The instructions need to be created as per user requests. User interaction commands are analyzed by the LLM model, which generates realistic and contextually relevant dialogues in VR. This enables VR training sessions to be more immersive and effective \cite{tang2025llm}. This integration allows trainees to engage in simulated environments that closely mimic real-world scenarios, enhancing their ability to learn and retain complex skills. For instance, LLM-driven chatbots can create dynamic training scenarios where workers practice communication and problem-solving skills in a controlled yet realistic setting \cite{li2024exploring}. To enhance the applicability of LLMs in VR, training the LLM model for specific use cases requires large amounts of data. Various methods are introduced for generating synthetic data by combining different skills derived from text, showcasing the potential of LLMs in creating diverse datasets \cite{park2025instruct}. This improves the applicability of LLMs for domain-related tasks. These studies collectively underscore the transformative impact of LLMs in VR, driving innovation and efficiency. 

Based on existing studies, users can effectively learn tasks in a virtual environment and create the virtual environment efficiently. LLMs have been utilized in various VR applications, such as code generation, VR scene creation, and collaborative design. Integrating LLMs with VR can simplify VR application development. Despite these advancements, the use of LLMs for generating the instructions autonomously in VR to train workers has not been thoroughly studied. In this paper, we propose a method to generate virtual instructions for worker training to perform assembly task using LLMs. 

\section{Generating Virtual Instruction using LLM}
\begin{figure}[H]
    \centering
    \includegraphics[width=0.49\textwidth]{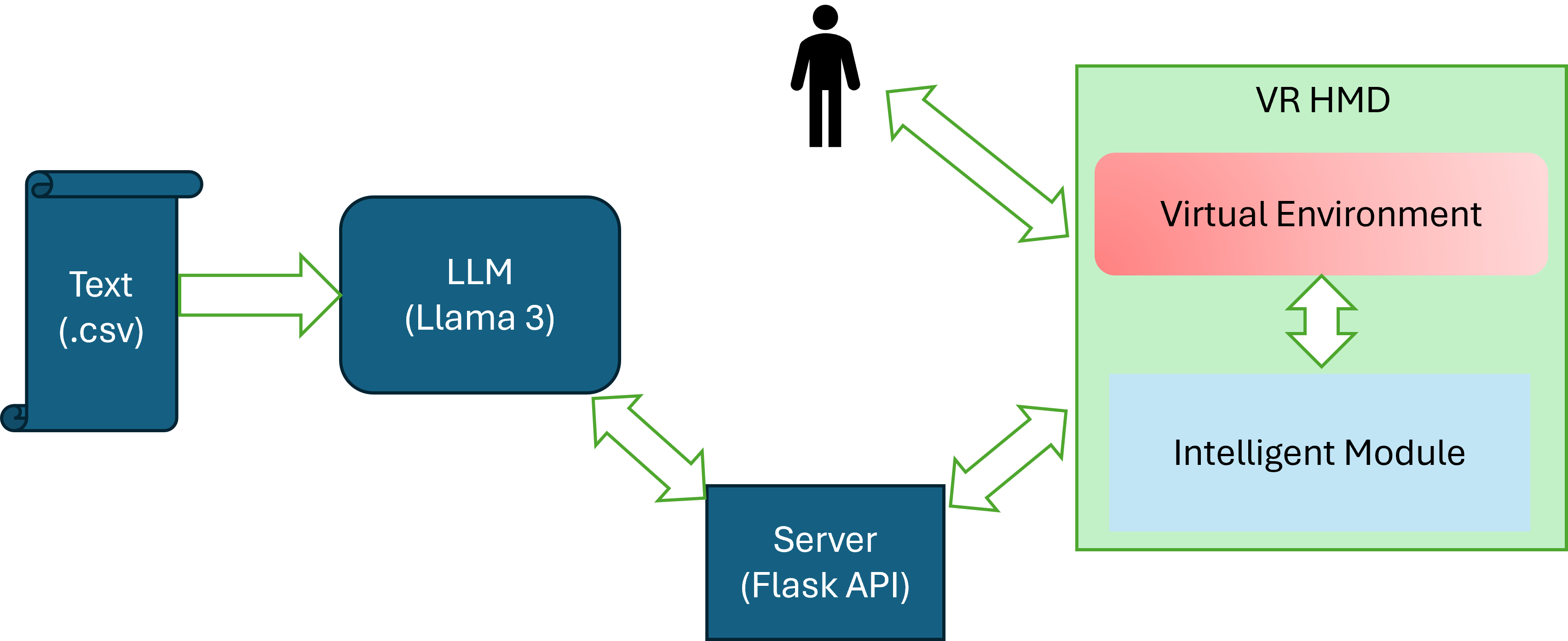}
    \caption{The architecture of virtual instruction generation.}\label{framework}
\end{figure}

This section discusses the generation of virtual instructions using LLMs. The system architecture of the proposed method is illustrated in Figure \ref{framework}, detailing the process of generating virtual instructions from text. The architecture comprises two main modules: the LLM module and the intelligent module. Communication between these modules is facilitated by the flask server which used TCP/IP protocol to establish communication between LLM module and Intelligent module. The text information is sent to the LLM module to extract relevant information from the text. The extracted information is then transferred to the intelligent module to generate the virtual instructions. These generated instructions are provided to the user for explaining the task.

\subsection{LLM Module}
\begin{figure}[H]
    \centering
    \includegraphics[width=0.49\textwidth]{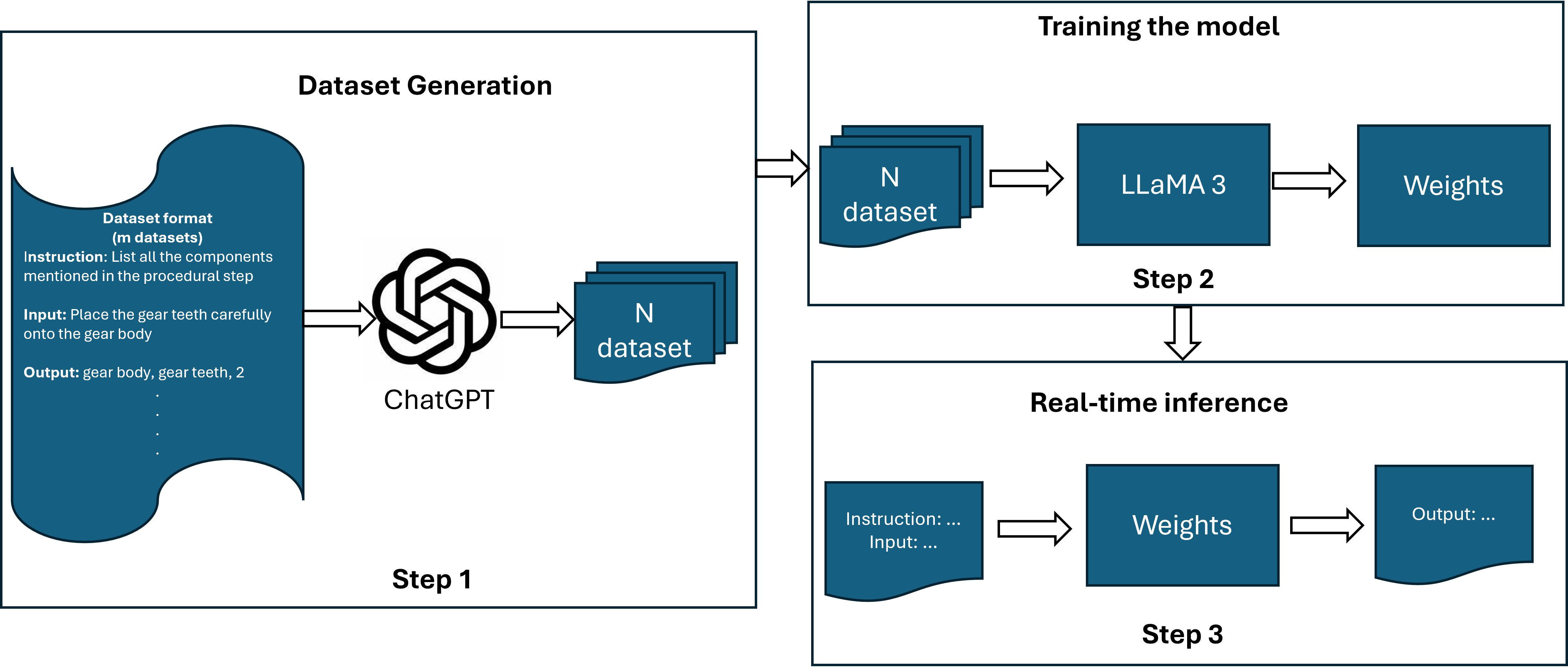}
    \caption{Steps involved in training LLM for domain specific task.}\label{llm training}
\end{figure}

LLMs often struggle with domain-specific accuracy due to limited exposure to specialized terminology and context during training. This can result in incorrect or overly general responses, reducing their reliability in expert-specific applications. The steps used to train the LLM for a specific domain are illustrated in Figure \ref{llm training}. The preparation of the LLM module consists of three main steps: dataset generation, model training, and real-time inference, as explained below.

\subsection{Dataset Generation}
LLM models are trained on massive amounts of data to make them capable of understanding instructions and generating different content to perform a wide range of tasks. The performance of the model might not be optimal in specific domains. Fine-tuning will improve the model's performance in specific domains. In fine-tuning, the model parameters are adjusted to work in specific domains while maintaining original knowledge. To fine-tune the model, it should be trained on a specific domain dataset. Dataset generation is a time-consuming and complex task. Synthetic data generation is one of the effective methods to create datasets. In this paper, synthetic data is generated using ChatGPT to train on a specific domain. The initial dataset was created based on the assembly task manually in Supervised Fine-Tuning format (SFT). The SFT dataset is structured as a JSON list of dictionaries, each containing an "instruction," an "input" (task-specific data), and a corresponding "output" representing the model's desired response. An example of the dataset is explained in Table \ref{table1}.

\begin{table}[ht]
\centering
\caption{Dataset format to train the LLM model} \label{table1}
\begin{tabularx}{0.49\textwidth}{|l|X|}
\hline
\textbf{Instruction} & List all the components mentioned in the procedural step. \\
\hline
\textbf{Input} & You need to fasten the base onto the assembled components and check that the four holes in the base align perfectly with the small screws. \\
\hline
\textbf{Output} & small screws, base, 1 \\
\hline
\end{tabularx}
\end{table}

We selected four assembly methods to fine-tune the LLM model: pneumatic cylinder assembly, diesel engine assembly, and two plugs in the hole assembly. These product assemblies require different steps and components. These datasets were generated using ChatGPT with the assistance of a small number of manually generated data (size = 60). Then, careful prompt engineering and filtering were used to ensure quality and diversity. We generated 420 datasets to fine-tune the model. The model was fine-tuned with three types of information: instruction, input, and output. The instruction assists the model in generating the output from the input. The input is assembly text instruction, and the output is the corresponding components in the assembly step and the number of similar components to be assembled.

\subsection{Training the Model}
\begin{figure}[H]
    \centering
    \includegraphics[width=0.49\textwidth]{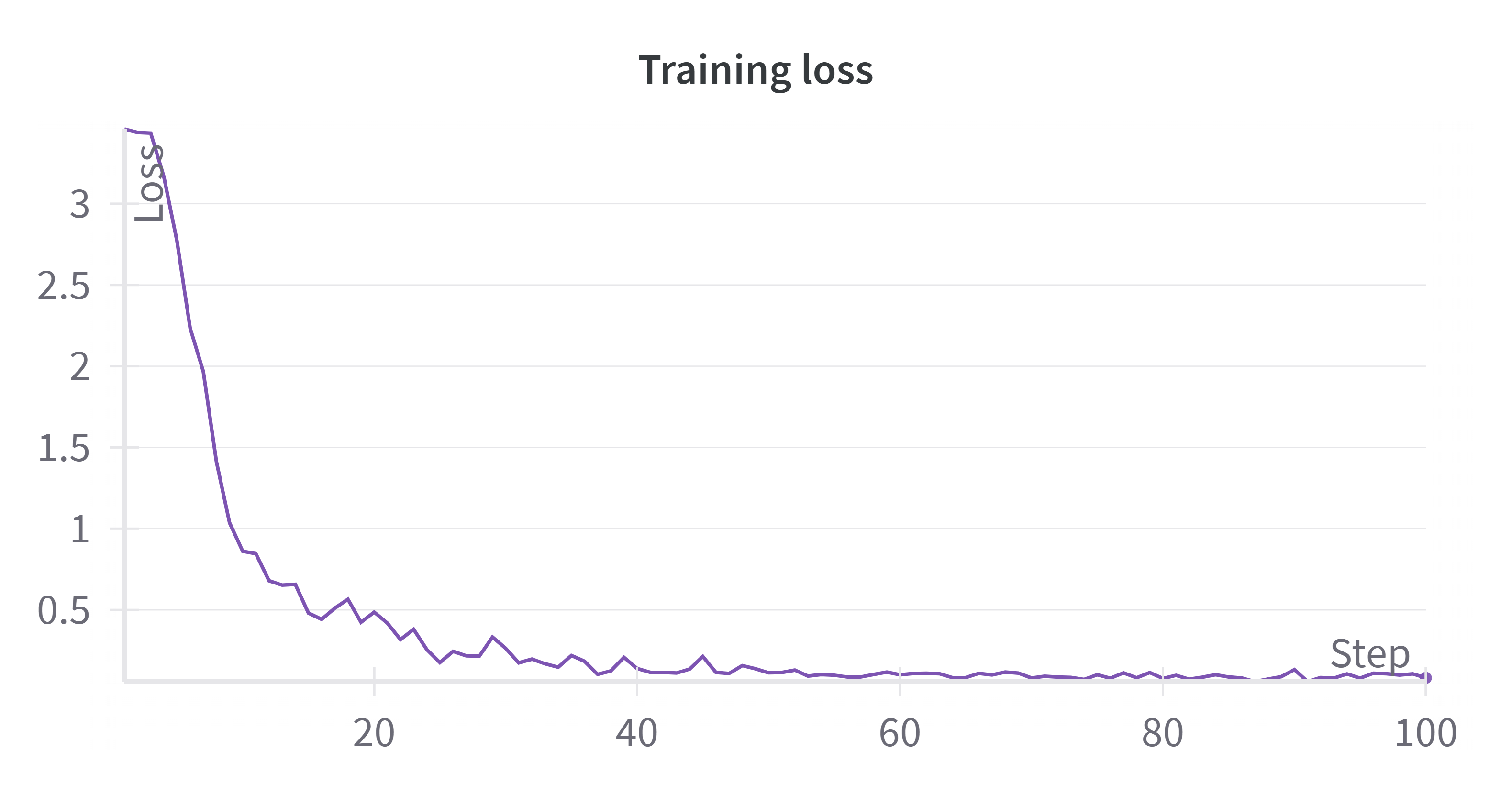}
    \caption{Training loss against the step.}\label{training loss}
\end{figure}

We fine-tuned the LLM model to extract information from text. Llama 3.0 is a Large Language Model that is faster and more efficient at inference compared to other LLM models. In this method, we used the Llama 3 model with 8 billion $(Llama3_8b)$ parameters for text-to-text generation to extract relevant information from text. We fine-tuned the 4-bit quantized $LLaMA-3 8B$  model using parameter-efficient fine-tuning (PEFT) via LoRA. The model was adapted on custom data with a maximum input sequence length of 2048 tokens. The LoRA configuration included a rank of 16, a scaling factor (LoRA alpha) of 16, and no dropout. Training was performed with an effective batch size of 8 (batch size 2 with gradient accumulation steps 4), using an AdamW 8-bit optimizer, a learning rate of 2e-4, linear learning rate decay, and weight decay of 0.01. We applied 5 warmup steps and trained for 100 total optimization steps. The training output was saved locally for inference. The training result is shown in figure \ref{training loss}, which explains the stability and minimal loss in the training process. We evaluated the fine-tuned model with test data, and the model was able to produce output effectively.

\subsection{Real-time Inference}
For inferencing, assembly steps are stored in the text file. This text input is input to the trained model to predict the information. This information is sent to the intelligent module through the TCP/IP protocol. The intelligent module generates the instructions based on the information. 

\subsection{Intelligent Module}
\begin{figure}[H]
    \centering
    \includegraphics[width=0.49\textwidth]{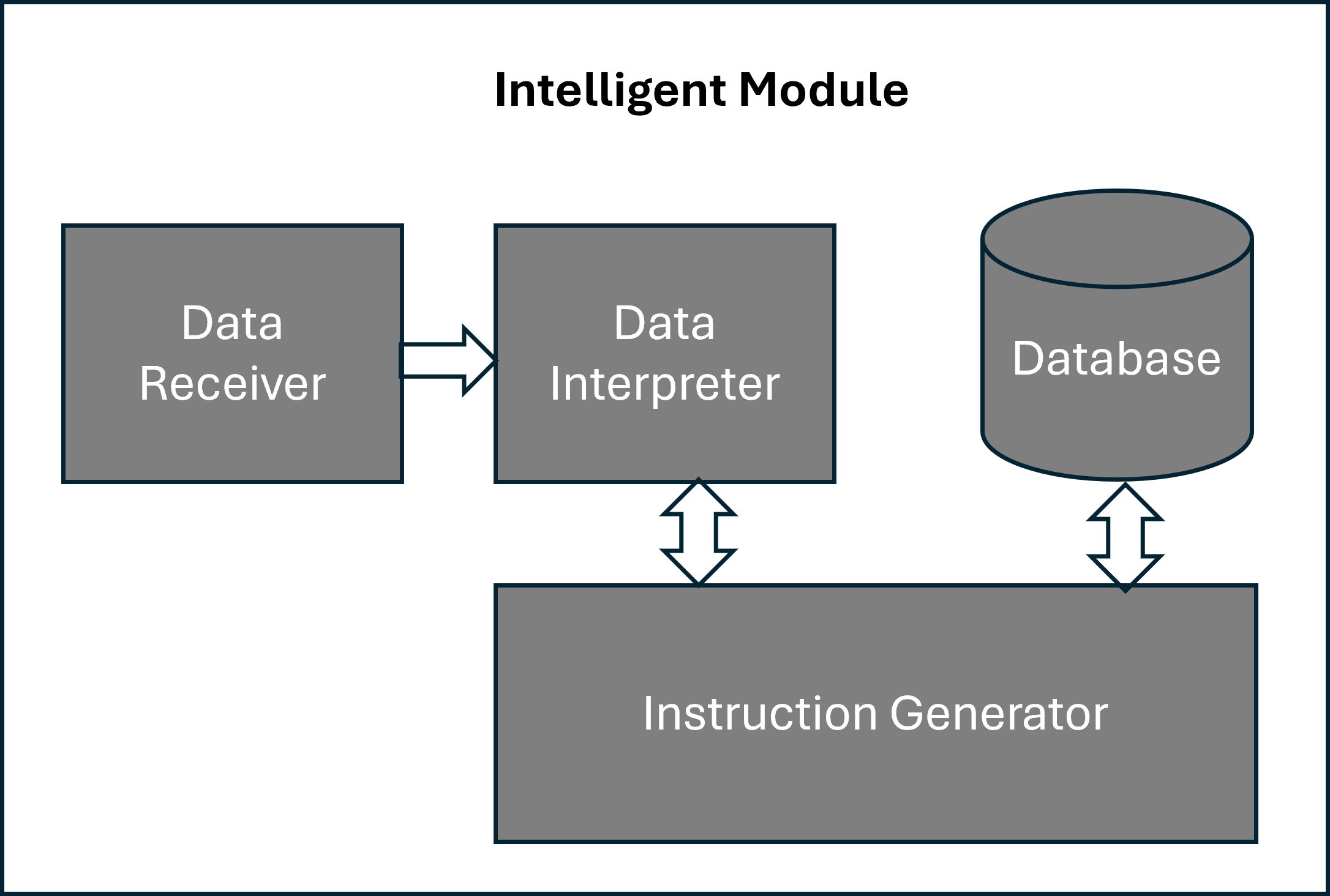}
    \caption{Components in intelligent module.}\label{intelligent module}
\end{figure}

Text information is sent to the LLM module, which extracts relevant information from the text. This information is then transferred to the intelligent system to generate virtual instructions. The intelligent system consists of a data receiver, data interpreter, a database , and an instruction generator module, which is shown in figure \ref{intelligent module}. The LLM module sends the output in a specific format that can be easily interpreted by the intelligent module. The output format of the LLM consists of the predecessor component name, followed by the successor component name, and then the number of successor components that need to be assembled. Data receivers receive the information from the LLM module through the server. This information is interpreted by the data interpreter, which identifies the precedence and successor component details, and the number of components that need to be assembled. This information is then transferred to the instruction generator. The instruction generator searches the database and activates the corresponding components in the virtual environment. Following that, the instruction generator generates virtual instructions and renders them in the virtual environment. The pseudocode  for instruction generation is explained in algorithm \ref{alg 1}.

\begin{algorithm}
\caption{Virtual Instruction Generation} \label{alg 1}
\begin{algorithmic}[1]
\State \textbf{Input:} Components name $[a, b]$, Number of components $n$
\State \textbf{Output:} Virtual instruction generation

\State Extract information from the input
\State Find relevant components $[A]$ in the database from $M$ components

\For{$i = 1$ to $M$}
    \If{$A_i == a$}
        \State Activate $A_i$
        \State Change the color of the component
    \ElsIf{$A_i == b$}
        \State Activate $A_i$
        \State Change the color of the component
    \Else
        \State Continue
    \EndIf
\EndFor

\State Generate the animation

\For{$i = 1$ to $M$}
    \If{$A_i == $\texttt{ a +\_+ b}}
        \State Activate $A_i$
        \State Create animation
        \State \textbf{break}
    \Else
        \State Continue
    \EndIf
\EndFor

\end{algorithmic}
\end{algorithm}

The dataset was created in the Unity environment, and all components were imported into the project using identical names. The database contains all relevant assembly components, imported as CAD models. These include both individual components and assembled components. In the case of assembled components, the parts that need to be assembled are saved under the name of the successor component. TThis distinct name helps the intelligent module identify the correct component from the database and generate the corresponding instruction. The data interpreter processes the received input and passes the interpreted information to the instruction generator. Based on this data, the instruction generator extracts the relevant models from the database. It then generates training instructions by changing the color of the components and creating animations that demonstrate how to assemble them, as explained in algorithm \ref{alg 1}. The intelligent module not only generates instructions for the next step but also deactivates previously generated instructions to avoid confusion during training. In the database, each CAD model is stored with a distinct name. The instruction generator accesses the material properties of the virtual components and changes their color to green. For animation creation, the intelligent module identifies the component name using a combination of the precedence and successor names and generates the animation by moving the successor component relative to the precedence component.

\section{Evaluation}
The proposed method evaluated in pneumatic cylinder assembly task. Assembly involves workers putting together a series of components to produce a final product. This task sequence is sent to the LLM model, which processes the instructions one by one from the text message of the assembly sequence and generates the output. The output includes details about the components involved in the current assembly step and the quantity of each component (e.g., number of screws or nuts). This information is then sent to intelligent module for generating the training instructions. 

\begin{figure}[H]
    \centering
    \includegraphics[width=0.49\textwidth]{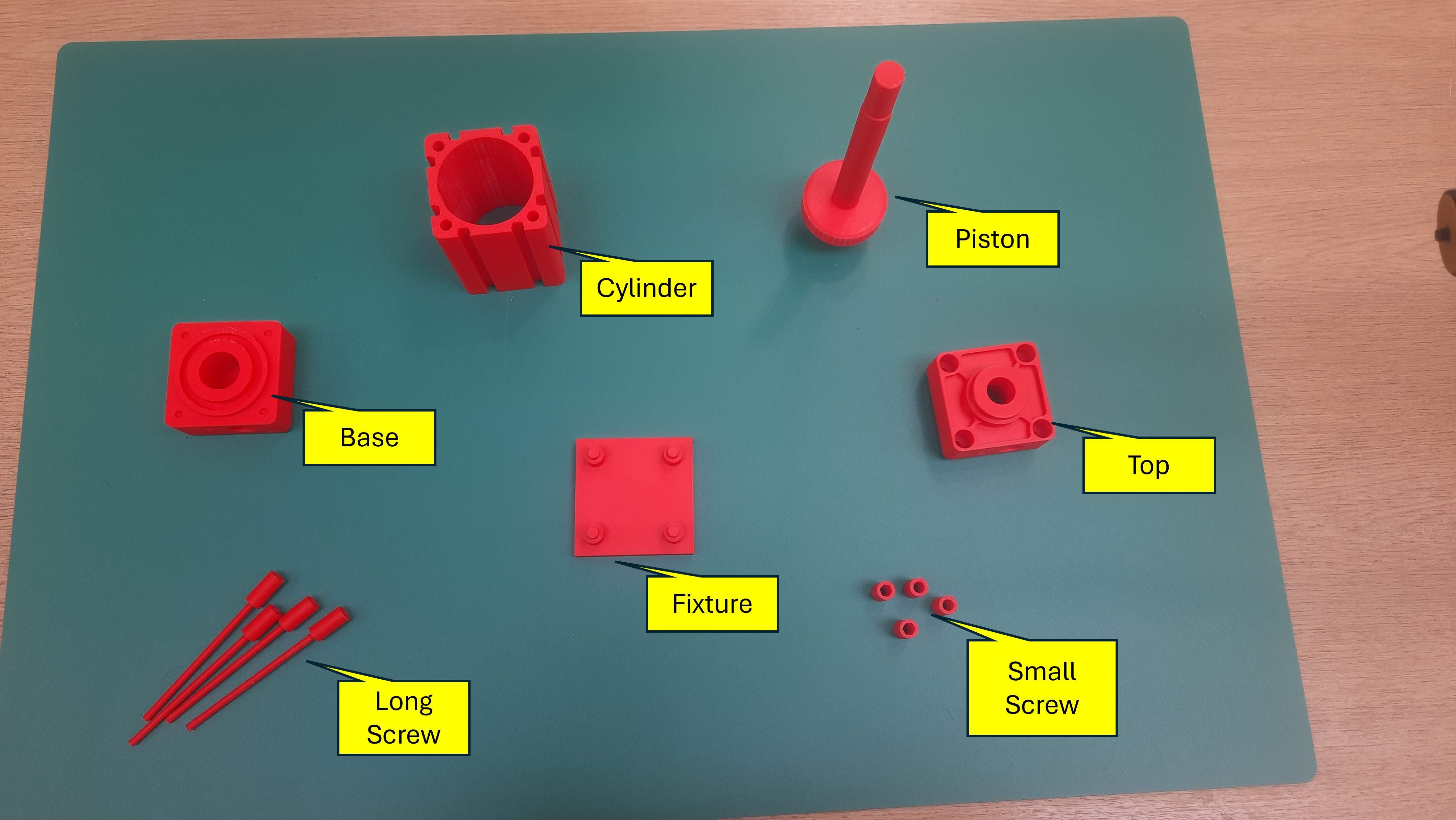}
    \caption{List of assembly components}\label{ASSEMBLY DETAILS}
\end{figure}

\begin{figure}[H]
    \centering
    \includegraphics[width=0.49\textwidth]{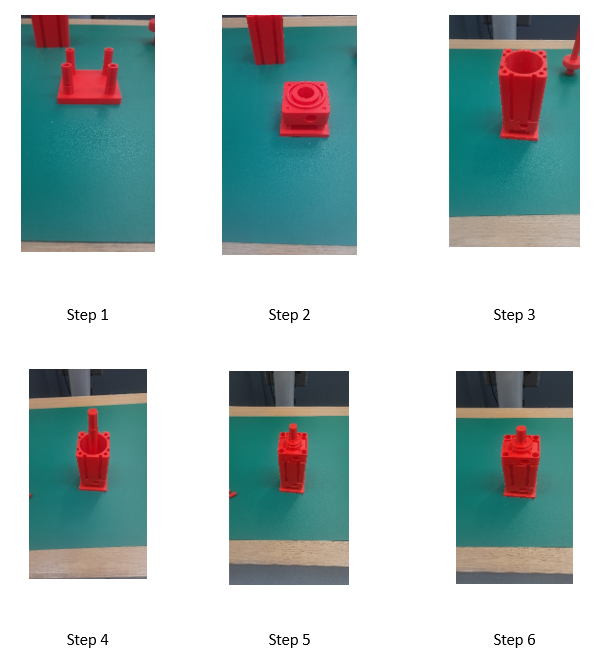}
    \caption{Details of the assembly step}\label{ASSEMBLY steps}
\end{figure}

We evaluated the proposed method using a pneumatic cylinder assembly. The pneumatic cylinder consists of the following components: fixture, base, cylinder, piston, top, small screw, and large screw, as shown in Figure \ref{ASSEMBLY DETAILS}. The assembly steps are explained in Figure \ref{ASSEMBLY steps}.

\begin{figure}[H]
    \centering
    \begin{subfigure}[b]{0.24\textwidth}
        \centering
        \includegraphics[width=\textwidth]{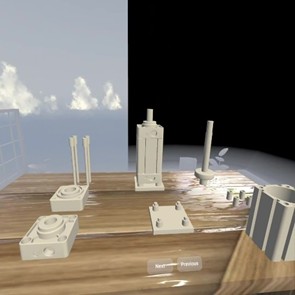}
        \caption{Before generating instruction}
        \label{fig:assembly1}
    \end{subfigure}
    \hfill
    \begin{subfigure}[b]{0.24\textwidth}
        \centering
        \includegraphics[width=\textwidth]{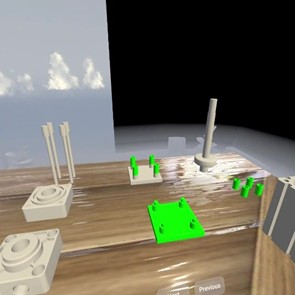}
        \caption{After generating instruction}
        \label{fig:assembly2}
    \end{subfigure}
    \caption{Generated assembly instruction}
    \label{fig:assembly}
\end{figure}

We developed the VR application using Unity, a game engine widely used for VR development. The Meta Quest 3 (MQ3) was used as the VR device, allowing users to experience the virtual environment. First, the VR application was created in Unity and then deployed on the MQ3. The application features a virtual environment with pneumatic cylinder components. The environment includes two buttons: "Next" and "Previous." Users access instructions by interacting with these buttons. When a button is clicked, the corresponding text instruction is sent to the LLM module, which extracts the relevant information. This information is then shared with the intelligent module to generate the virtual instruction, as explained in Section 3. The generated instruction is shown in figure \ref{fig:assembly}. The generated instruction is indicated as green color which is shown in figure \ref{fig:assembly2}. Similarly, users interact with the LLM module by selecting buttons to learn each task. The proposed method is capable of generating instructions effectively and accurately.

\section{Conclusion}
This paper proposed a method to automatically generate virtual instructions from textual input. By leveraging the capabilities of LLMs and an intelligent module, the system effectively extracts task-relevant information and transforms it into immersive, animated VR instructions. This method significantly reduces the time and expertise required to develop VR training content, making it more scalable and adaptable to industrial needs. The proposed system was implemented and evaluated in the context of pneumatic cylinder assembly using Unity and deployed on the Meta Quest 3 headset. The VR environment allowed users to interact with virtual components and receive step-by-step guidance through intuitive controls. The communication between the LLM and intelligent modules, facilitated via HTTP, enabled real-time instruction generation based on user input. Experimental results demonstrated that the system could accurately interpret assembly instructions from text and generate corresponding virtual guidance, enhancing the learning experience. Overall, this approach offers a promising direction for automating and scaling VR training solutions across various industrial domains. In the future, we plan to evaluate the proposed method through a user study to analyze the effectiveness of the generated instructions.

\end{document}